\documentclass{article}

\PassOptionsToPackage{numbers}{natbib}
\usepackage[preprint]{nips_2018}

\usepackage[utf8]{inputenc} 
\usepackage[T1]{fontenc}    
\usepackage{hyperref}       
\usepackage{url}            
\usepackage{booktabs}       
\usepackage{amsfonts}       
\usepackage{nicefrac}       
\usepackage{microtype}      
\usepackage{textcomp}
\usepackage{graphicx}
\usepackage{caption}
\usepackage{floatrow}

\title{Real-time 2D Multi-Person Pose Estimation on CPU: Lightweight OpenPose}

\author{
  Daniil Osokin \\
  Intel\\
  \texttt{daniil.osokin@intel.com} \\
}

\begin{document}
\bibliographystyle{plainnat}

\maketitle

\begin{abstract}
  In this work we adapt multi-person pose estimation architecture to use it on edge devices. We follow the bottom-up approach from OpenPose \cite{cao2017realtime}, the winner of COCO 2016 Keypoints Challenge, because of its decent quality and robustness to number of people inside the frame. With proposed network design and optimized post-processing code the full solution runs at 28 frames per second (fps)  on  Intel\textsuperscript{\textregistered{}} NUC 6i7KYB mini PC and 26 fps on Core i7-6850K CPU. The network model has 4.1M parameters and 9 billions floating-point operations (GFLOPs) complexity, which is just $\sim$15\% of the baseline 2-stage OpenPose with almost the same quality. The code and model are available as a part of Intel\textsuperscript{\textregistered{}} OpenVINO\textsuperscript{TM} Toolkit.
\end{abstract}

\section{Introduction}

Multi-person pose estimation is an important task and may be used in different domains, such as action recognition, motion capture, sports, etc. The task is to predict a pose skeleton for every person in an image. The skeleton consists of keypoints (or joints): ankles, knees, hips, elbows, etc.

Human pose estimation accuracy was greatly improved with the help of convolutional neural networks (CNNs) \cite{he2017maskrcnn}, \cite{fang2017rmpe}, \cite{xiao2018simple}. However, there is a little research on compact, yet efficient pose estimation methods. In \cite{maskrcnn2go} authors show a simplified Mask R-CNN keypoint detector demo on a mobile phone, running at 10 fps, however neither implementation details nor accuracy characteristics were provided. We have also found the open-source repository \cite{ildoonet} with human pose estimation network. Author reported inference speed of 4.2 fps on 2.8GHz Quad-core CPU and 10 fps on Jetson TX2 board.

In our work we optimize the popular method OpenPose and show how modern design techniques of CNNs can be used for pose estimation task. As a result, our solution runs at:
\begin{itemize}
  \item 28 fps on mini PC  Intel\textsuperscript{\textregistered{}} NUC, which consumes little power and has 45 watt CPU TDP.
  \item 26 fps on a usual CPU without the need of a graphic card.
\end{itemize}
The accuracy of the optimized version nearly matches the baseline: Average Precision (AP) drop is less than 1\%.

\section{Related Work}

Multi-person pose estimation problem can usually be approached in two ways. The first one, called {\it top-down}, applies a person detector and then runs a pose estimation algorithm per every detected person. So pose estimation problem is decoupled into two subproblems, and the state-of-the-art achievements from both areas can be utilized. The inference speed of this approach strongly depends on number of detected people inside the image.

The second one, called {\it bottom-up}, more robust to the number of people. At first all keypoints are detected in a given image, then they are grouped by human instances. Such approach usually faster than the previous, since it finds keypoints once and does not rerun pose estimation for each person.

In \cite{kocabas18prn} authors proposed the fastest method to date with state-of-the-art quality among bottom-up methods, which runs 23 fps on a single GTX 1080 Ti graphic card for an image with 3 persons. They note, that performance will degrade to 15 fps for image with 20 persons. We based our work on the popular bottom-up method OpenPose, it has almost invariant to number of people inference time.

\section{Analysis of the Original OpenPose}

\subsection{Inference Pipeline}
Similar to all bottom-up methods, OpenPose pipeline consist of two parts:
\begin{itemize}
  \item Inference of Neural Network to provide two tensors: keypoint heatmaps and their pairwise relations (part affinity fields, pafs). This output is downsampled 8 times.
  \item Grouping keypoints by person instances. It includes upsampling tensors to original image size, keypoints extraction at the heatmaps peaks and their grouping by instances.
\end{itemize}

\begin{figure}[h]\CenterFloatBoxes
\begin{floatrow}
\ffigbox[\FBwidth]
  {\includegraphics[height=4.2cm]{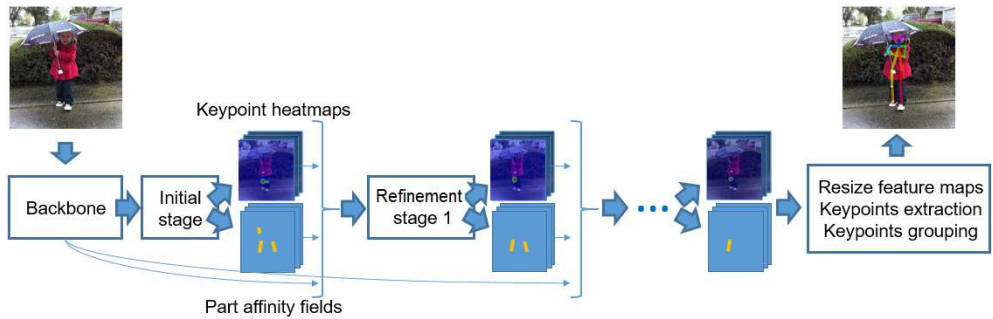}}
  {\caption{OpenPose pipeline.}
   \label{OpenPosePipeline}}
  \end{floatrow}
 \end{figure}

The network first extracts features, then performs initial estimation of heatmaps and pafs, after that 5 refinement stages are performed. It is able to find 18 types of keypoints. Then grouping procedure searches the best  pair (by affinity) for each keypoint, from the predefined list of keypoint pairs, e.g. {\it left elbow} and {\it left wrist}, {\it right hip} and {\it right knee}, {\it left eye} and {\it left ear}, and so on, 19 pairs overall. The pipeline is illustrated in Fig.~\ref{OpenPosePipeline}. During inference, input image is resized to match network input size by height, the width is scaled to preserve image aspect ratio, then padded to the multiple of 8.

\subsection{Complexity Analysis}

The original implementation uses VGG-19 backbone \cite{vgg} cut to \textit{conv4\_2} layer as a features extractor. Then two extra convolutional layers \textit{conv4\_3} and \textit{conv4\_4} are added. After that initial and 5 refinement stages are made.

Each stage consists of two parallel branches: one for heatmaps estimation and one for pafs. The two branches have the same design, shown in Table~\ref{stages_design}. We set network input resolution to 368x368 in our comparison and use the same COCO validation subset as in original paper, single scale testing is performed. The test CPU is Intel\textsuperscript{\textregistered{}} Core\textsuperscript{TM} i7-6850K, 3.6GHz. Table~\ref{gmac_vs_accuracy_baseline} shows the trade-off between accuracy and number of refinement stages.

It can be seen, that the latter stages give less improvement per GFLOPs, so for the optimized version we will keep only the first two stages: the initial stage and a single refinement stage.

The profile for the post-processing part is summarized in the Table~\ref{grouping_profile}. It was obtained by running the code, which was written in C++ with OpenCV \cite{opencv_library}. Despite the grouping itself is lightweight, other parts are subject to optimization.
\floatsetup[table]{style=Plaintop}
\begin{table}[ht]
  \begin{floatrow}
    \ttabbox
    {\caption{OpenPose stages design. Each stage has 2 parallel branches (single is shown).}
    \label{stages_design}}
    {  \begin{tabular}{cc}
    \toprule
    Initial & Refinement \\
    \midrule
    conv 3x3/128 & conv 7x7/128 \\
    conv 3x3/128 & conv 7x7/128 \\
    conv 3x3/128 & conv 7x7/128 \\
    conv 1x1/512 & conv 7x7/128 \\
                 & conv 7x7/128 \\
                 & conv 1x1/128 \\
    \bottomrule
  \end{tabular}}
    \ttabbox
    {\caption{Accuracy vs. Complexity of OpenPose on COCO validation set.}
    \label{gmac_vs_accuracy_baseline}}
    {  \begin{tabular}{lccc}
    \toprule
                           {}  & {AP, \%} & {GFLOPs} & {GFLOPs} \\
                            {} &  {}      & {}       & {total} \\
    \midrule
    Backbone                 & n/a  & 37.8 & 37.8  \\
    conv4\_3                 & n/a  & 2.5  & 40.3 \\
    conv4\_4                 & n/a  & 0.6  & 40.9  \\
    Initial stage            & 35.5 & 2.2  & 43.1 \\
    Refinement stage 1       & 43.4 & 18.6 & 61.7 \\
    Refinement stage 2       & 46.2 & 18.6 & 80.3 \\
    Refinement stage 3       & 47.4 & 18.6 & 98.9 \\
    Refinement stage 4       & 48.1 & 18.6 & 117.5 \\
    Refinement stage 5       & 48.6 & 18.6 & 136.1 \\
    \bottomrule
  \end{tabular}}
  \end{floatrow}
\end{table}

\begin{table}[ht]
  \begin{floatrow}
    \ttabbox
    {\caption{Initial performance of post-processing and grouping.}
     \label{grouping_profile}}
    {
  \begin{tabular}{lcccc}
    \toprule
    & Resize feature maps & Extract keypoints & Group keypoints & Total \\
    \midrule
    Fps & 10.5              & 1.81            & 454             & \textbf{1.54} \\
    \bottomrule
  \end{tabular}
  }
  \end{floatrow}
\end{table}

\section{Optimization}

\subsection{Network Design}

All experiments were performed with the default training parameters form the original paper, and we used the COCO dataset \cite{mscoco} to train on. As pointed above, we keep only initial and first refinement stage. However, the rest stages can provide regularizing effect, so the final network was retrained with additional stages, but the first two are used. Such procedure gives $\sim$1\% AP improvement.

\subsubsection{Lightweight Backbone}
Since time when VGG nets were proposed, few lightweight network topologies with similar or even better classification accuracy were designed \cite{hong2016pvanet},  \cite{mobilenet_v1}, \cite{mobilenet_v2}. We evaluated networks from MobileNet family to replace the VGG feature extractor and started from MobileNet v1.

\begin{figure}[ht]\CenterFloatBoxes
\begin{floatrow}
\ttabbox[\FBwidth]
{\caption{Lightweight backbone selection study (the initial and refinement stages have original OpenPose design).}
    \label{ablation_backbone_design}}
    {  \begin{tabular}{lcc}
    \toprule
    & AP, \% & GFLOPs \\
    \midrule
    MobileNet v1          & 37.9 & 23.3 \\
    {\it (cut to conv4\_1)} & & \\
    Dilated MobileNet v1  & 42.8 & 27.7 \\
    {\it (cut to conv5\_5)} & & \\
    Dilated MobileNet v1  & 43.2 & 31.3  \\
    {\it (cut to conv5\_6)} & & \\
    Dilated MobileNet v2  & 39.6 & 27.2 \\
    {\it (cut to conv6\_3)} & & \\
    \bottomrule
  \end{tabular}}
\killfloatstyle
\ffigbox[]
  {\includegraphics[height=2.8cm]{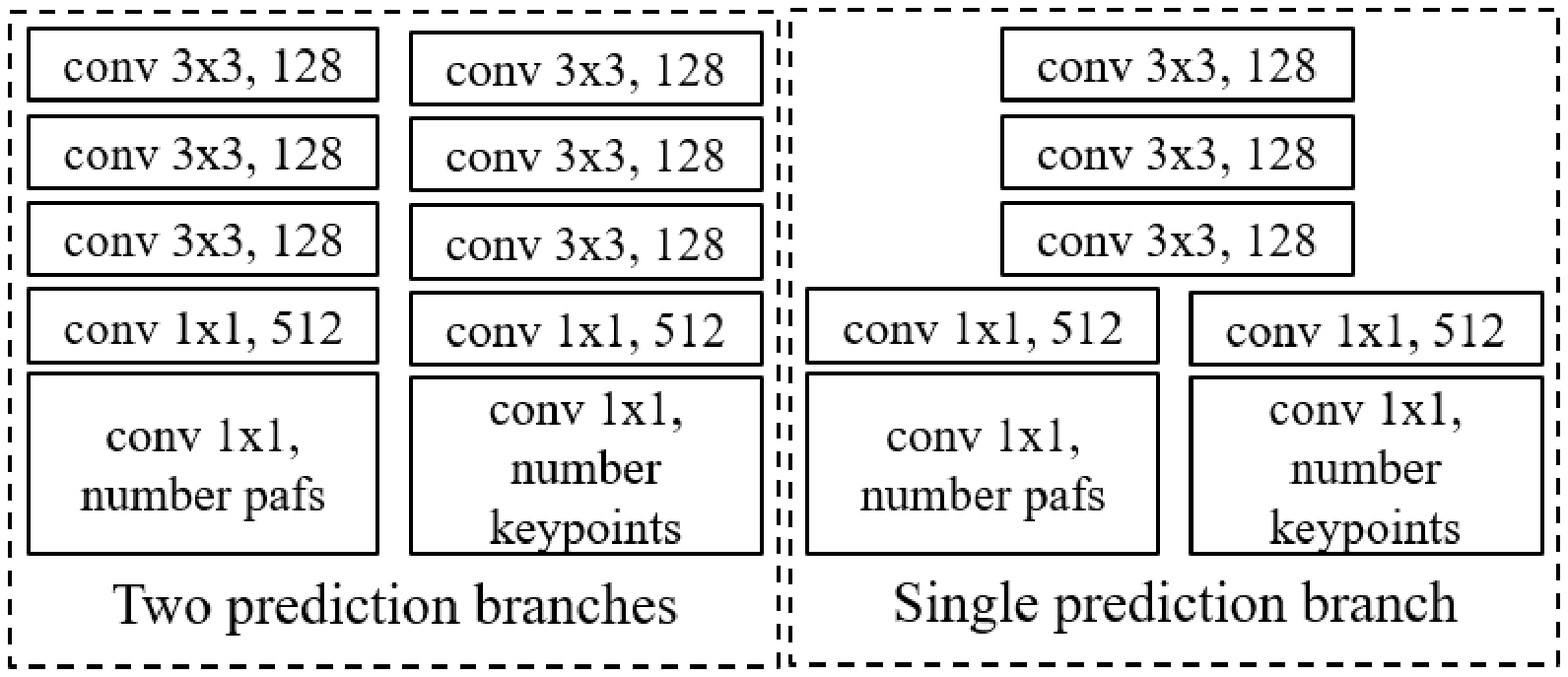}}
  {\caption{Original two prediction branches and proposed single prediction branch for the initial stage. We also apply this scheme for the refinement stage.}
   \label{single_branch}}
  \end{floatrow}
 \end{figure}

In a naive way, if we keep all layers till deepest, which matched output tensor resolution, it leads to significant accuracy drop. This might be due to shallowness and weak feature representation. To save spatial resolution and reuse backbone weights we use dilated convolution \cite{drn}. Stride of {\it conv4\_2/dw} layer was removed and dilation parameter value was set to 2 for succeeding {\it conv5\_1/dw} layer to preserve receptive field. So we use all layers till {\it conv5\_5} block. Addition of {\it conv5\_6} block improves the accuracy, but at cost of performance. We also tried more lightweight backbone MobileNet v2, however it did not show good result, see Table~\ref{ablation_backbone_design}.

\subsubsection{Lightweight Refinement Stage}

To produce new estimation of keypoint heatmaps and pafs the refinement stage takes features from backbone, concatenated with previous estimation of keypoint heatmaps and pafs. Motivated by this fact we decided to share the most of computations between heatmaps and pafs and use {\it single prediction branch} in initial and refinement stage. We share all layers except the two last, which directly produce keypoint heatmaps and pafs, see Fig.~\ref{single_branch}.

\begin{figure}[ht]\CenterFloatBoxes
\begin{floatrow}
\ffigbox[5cm]
  {\includegraphics[height=3cm]{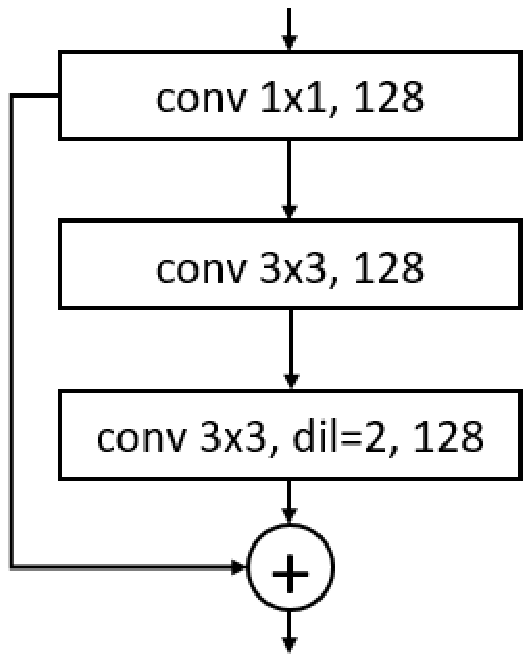}}
  {\caption{Design of convolutional block for replacement convolutions with 7x7 kernel size in refinement stage.}
   \label{conv7x7_substitution}}
\killfloatstyle
\ttabbox[\Xhsize]
    {\caption{Accuracy versus Complexity of proposed network on COCO validation set.}
    \label{gmac_vs_accuracy_final}}
    {  \begin{tabular}{lccc}
    \toprule
                             & AP, \% & GFLOPs & GFLOPs  \\
                             &        &        & total  \\
    \midrule
    Dilated MobileNet v1     & n/a  & 3.7 & 3.7 \\
    conv4\_3                 & n/a  & 0.3 & 4 \\
    conv4\_4                 & n/a  & 0.3 & 4.3 \\
    Initial stage            & 35   & 1.3 & 5.6 \\
    Refinement stage 1       & 41.4 & 3.4 & 9\\
    \hline
     2-stage network, retrained & & & \\
     with all refinement stages &\textbf{42.8} & n/a & \textbf{9} \\
    \bottomrule
  \end{tabular}}
  \end{floatrow}
 \end{figure}

Then each convolution with 7x7 kernel size was replaced by a convolutional block with the same receptive field, to capture long-range spatial dependencies \cite{cpm}. We conducted series of experiments with this block design and observed that it's enough to have three consecutive convolutions with 1x1, 3x3, and 3x3 kernel size, the latter with dilation parameter equals to 2, to preserve initial receptive field. Because the network became deeper, we added residual connection \cite{he2016resnet} for each such block. The final design visualized in Fig.~\ref{conv7x7_substitution}, it has $\sim$2.5 times less complexity than convolution with 7x7 kernel. We also replaced conv4\_3 with 3 depthwise separable convolutions, channels number was reduced from 256 to 128. The complexity and accuracy of the proposed network design are shown in the Table~\ref{gmac_vs_accuracy_final}.

\subsection{Fast Post-processing}

We profiled the code and removed extra memory allocations, parallelized keypoints extraction with OpenCV's routine. This made code significantly faster, and the last bottleneck was the resize feature maps to the input image size.

We decided to skip the resize step and performed grouping directly on network output, but accuracy dropped significantly. Thus step with upsampling feature maps cannot be avoided, but it is not necessary to do it to input image size. Our experiments shown, that with upsample factor 8 the accuracy is the same, as if resize to input image size. We used upsample factor 4 for the demo purposes.

\subsection{Inference}

For the network inference we use the  Intel\textsuperscript{\textregistered{}} OpenVINO\textsuperscript{TM} Toolkit R4 \cite{openvino_toolkit}, which provides optimized inference across different hardware, such as CPU, GPU, FPGA, etc. Final performance numbers are shown in the Table~\ref{final_results}, they were measured for a challenging video with more than 20 estimated poses.

\begin{table}[h]
  \begin{floatrow}[1]
    \ttabbox[\Xhsize]
    {\caption{Final inference fps for a video with more than 20 estimated poses. Numbers in braces are network inference and post-processing fps.}
    \label{final_results}}
    {
  \begin{tabular}{lcc}
    \toprule
             & NUC              & CPU \\
    \midrule
    Baseline & 1.17 (3.92/1.66) & 0.95 (2.47/1.54) \\
    Proposed & \textbf{28} (33/160)      & \textbf{26} (33/125)      \\
    \bottomrule
  \end{tabular}
    }
  \end{floatrow}
\end{table}

We used two devices: Intel NUC6i7KYB, which performed inference on the integrated GPU Iris Pro Graphics P580 in half-precision floating-point format (FP16), and 6-core Core i7-6850K CPU, which performed inference in single-precision floating-point format (FP32). Network input size was set to 456x256, which is similar to 368x368, but with 16:9 aspect ratio, suitable for processing video streams.

\section{Conclusion}

In this work, we approached the problem of human pose estimation network, suitable for real-time performance on edge devices. We proposed the solution, based on OpenPose method, with heavily optimized network design and post-processing code. The accuracy versus network complexity ratio was increased in more than 6.5 times due to the use of dilated MobileNet v1 feature extractor with depthwise separable convolutions and design of lightweight refinement stage with residual connections. The network can be downloaded as a part of the \href{https://software.intel.com/en-us/openvino-toolkit/choose-download}{OpenVINO Toolkit} under the name \textit{human-pose-estimation-0001}. The network description is available in the  \href{https://github.com/opencv/open_model_zoo/tree/2018/intel_models}{Open Model Zoo} repository.

The full solution runs in real time on a usual CPU, as well as on NUC mini PC and closely matches accuracy of the baseline 2-stage network. Some techniques may further improve performance and accuracy, such as quantization, pruning, knowledge distillation. We left them for the future research.


{\small
\bibliography{hpe_realtime_cpu_arxiv}}

\begin{thebibliography}{17}
\providecommand{\natexlab}[1]{#1}
\providecommand{\url}[1]{\texttt{#1}}
\expandafter\ifx\csname urlstyle\endcsname\relax
  \providecommand{\doi}[1]{doi: #1}\else
  \providecommand{\doi}{doi: \begingroup \urlstyle{rm}\Url}\fi

\bibitem[ope()]{openvino_toolkit}
{OpenVINO Toolkit}.
\newblock \url{https://software.intel.com/en-us/openvino-toolkit}.

\bibitem[Bradski(2000)]{opencv_library}
G.~Bradski.
\newblock {The OpenCV Library}.
\newblock \emph{Dr. Dobb's Journal of Software Tools}, 2000.

\bibitem[Cao et~al.(2017)Cao, Simon, Wei, and Sheikh]{cao2017realtime}
Z.~Cao, T.~Simon, S.~Wei, and Y.~Sheikh.
\newblock {Realtime Multi-Person 2D Pose Estimation using Part Affinity
  Fields}.
\newblock In \emph{CVPR}, 2017.

\bibitem[Fang et~al.(2017)Fang, Xie, Tai, and Lu]{fang2017rmpe}
H.-S. Fang, S.~Xie, Y.-W. Tai, and C.~Lu.
\newblock {RMPE: Regional Multi-person Pose Estimation}.
\newblock In \emph{ICCV}, 2017.

\bibitem[He et~al.(2016)He, Zhang, Ren, and Sun]{he2016resnet}
K.~He, X.~Zhang, S.~Ren, and J.~Sun.
\newblock {Deep Residual Learning for Image Recognition}.
\newblock In \emph{CVPR}, 2016.

\bibitem[He et~al.(2017)He, Gkioxari, Doll\'{a}r, and Girshick]{he2017maskrcnn}
K.~He, G.~Gkioxari, P.~Doll\'{a}r, and R.~Girshick.
\newblock {Mask R-CNN}.
\newblock In \emph{ICCV}, 2017.

\bibitem[Hong et~al.(2016)Hong, Roh, Kim, Cheon, and Park]{hong2016pvanet}
S.~Hong, B.~Roh, K.-H. Kim, Y.~Cheon, and M.~Park.
\newblock {PVANet: Lightweight Deep Neural Networks for Real-time Object
  Detection}.
\newblock In \emph{arXiv preprint arXiv:1611.08588}, 2016.

\bibitem[Howard et~al.(2017)Howard, Zhu, Chen, Kalenichenko, Wang, Weyand,
  Andreetto, and Adam]{mobilenet_v1}
A.~G. Howard, M.~Zhu, B.~Chen, D.~Kalenichenko, W.~Wang, T.~Weyand,
  M.~Andreetto, and H.~Adam.
\newblock {Mobilenets: Efficient convolutional neural networks for mobile
  vision applications}.
\newblock In \emph{arXiv preprint arXiv:1704.04861}, 2017.

\bibitem[Jindal and et~al.(2018)]{maskrcnn2go}
A.~Jindal and et~al.
\newblock Enabling full body ar with mask r-cnn2go.
\newblock In
  \emph{https://research.fb.com/enabling-full-body-ar-with-mask-r-cnn2go/},
  2018.

\bibitem[Kim()]{ildoonet}
I.~Kim.
\newblock tf-pose-estimation.
\newblock \url{https://github.com/ildoonet/tf-pose-estimation}.

\bibitem[Kocabas et~al.(2018)Kocabas, Karagoz, and Akbas]{kocabas18prn}
M.~Kocabas, S.~Karagoz, and E.~Akbas.
\newblock Multi{P}ose{N}et: Fast multi-person pose estimation using pose
  residual network.
\newblock In \emph{ECCV}, 2018.

\bibitem[Lin et~al.(2014)Lin, Maire, Belongie, Hays, Perona, Ramanan,
  Doll\'{a}r, and Zitnick]{mscoco}
T.-Y. Lin, M.~Maire, S.~Belongie, J.~Hays, P.~Perona, D.~Ramanan,
  P.~Doll\'{a}r, and C.~L. Zitnick.
\newblock {Microsoft COCO: common objects in context}.
\newblock In \emph{ECCV}, 2014.

\bibitem[Sandler et~al.(2018)Sandler, Howard, Zhu, Zhmoginov, and
  Chen]{mobilenet_v2}
M.~Sandler, A.~G. Howard, M.~Zhu, A.~Zhmoginov, and L.~Chen.
\newblock {MobileNetV2: Inverted Residuals and Linear Bottlenecks}.
\newblock In \emph{CVPR}, 2018.

\bibitem[Simonyan and Zisserman(2015)]{vgg}
K.~Simonyan and A.~Zisserman.
\newblock Very deep convolutional networks for large-scale image recognition.
\newblock In \emph{ICLR}, 2015.

\bibitem[Wei et~al.(2016)Wei, Ramakrishna, Kanade, and Sheikh]{cpm}
S.~Wei, V.~Ramakrishna, T.~Kanade, and Y.~Sheikh.
\newblock Convolutional pose machines.
\newblock In \emph{CVPR}, 2016.

\bibitem[Xiao et~al.(2018)Xiao, Wu, and Wei]{xiao2018simple}
B.~Xiao, H.~Wu, and Y.~Wei.
\newblock {Simple Baselines for Human Pose Estimation and Tracking}.
\newblock In \emph{ECCV}, 2018.

\bibitem[Yu et~al.(2017)Yu, Koltun, and Funkhouser]{drn}
F.~Yu, V.~Koltun, and T.~Funkhouser.
\newblock Dilated residual networks.
\newblock In \emph{CVPR}, 2017.

\end{thebibliography}

\end{document}